# Predicting Question Quality on StackOverflow with Neural Networks

**Mohammad Al-Ramahi**
Texas A&M University - San Antonio
mrahman1@tamusa.edu

**Izzat Alsmadi**
Texas A&M University - San Antonio
ialsmadi@tamusa.edu

**Abdullah Wahbeh**
Slippery Rock University of Pennsylvania
Abdullah.wahbeh@sru.edu

## Abstract

The wealth of information available through the Internet and social media is unprecedented. Within computing fields, websites such as Stack Overflow are considered important sources for users seeking solutions to their computing and programming issues. However, like other social media platforms, Stack Overflow contains a mixture of relevant and irrelevant information.

In this paper, we evaluated neural network models to predict the quality of questions on Stack Overflow, as an example of Question Answering (QA) communities. Our results demonstrate the effectiveness of neural network models compared to baseline machine learning models, achieving an accuracy of 80%. Furthermore, our findings indicate that the number of layers in the neural network model can significantly impact its performance.

## Introduction

The Internet provides a wealth of resources for knowledge seekers. One advantage of reaching out to information through the internet is related to the freshness of information and users' experience. Nowadays, question answering websites are becoming increasingly popular and used by many professionals to find answers to their problems (Hu et al. 2018). Such websites play crucial role in fulfilling the information need of users (Roy 2020). One popular question answering website is StackOverFlow. StackOverflow allows users to post their questions related to all their information technology, computing, and programming problems or issues. They can also read through a wealth of questions and answers to similar questions posted by other users.

StackOverflow host millions of questions and answers to technical problems where less experienced users are most likely to spend more time on the Website to find answers to their problems or even post new question (Öztürk 2022). In comparison with information in books or even on most other websites, information on StackOverflow is always tagged with dates and has most recent responses to similar questions. This is important as most of those problems discussed on StackOverflow are related to operating systems, software applications or programming libraries with certain versions or releases. In such cases, these problems can be very context dependent where users can check if they match their own context.

The quality of the content posted on question answering websites such as StackOverflow varies from high quality content to low quality content (Roy 2020). This means that some of the posted problems on these websites remain unsolved, which might affect users' experience. Even in cases where solutions could be found, there is no guarantee that these solutions are considered high quality solutions. Hu et al., (Hu et al. 2018) argued that low quality answers on question answering Websites can negatively affect users' experience with these services.

User experience in websites such as StackOverflow is critical since there may be no structured process to audit and authenticate experts and their responses to users' questions, yet in most cases users find expert responses very useful and relevant.

On question answering website, any user, regardless of his or her experience can respond back to questions (Tóth et al. 2019). As a result, this could end up with a lot of irrelevant and low-quality information. This is common in most social network websites especially as they get popular. Spammers from different categories and for different purposes try to inject their information and make such information visible and accessible to all users.

StackOverflow strive to maintain high quality of the content posted the community by either closing and/or deleting inappropriate posts (Tóth et al. 2019). Even with such approaches, question answering websites still have many challenges to address, these include but not limited to, dealing with large content, detection of unfaithful post, duplicate question removal, and the ability to find and locate best solution for specific question (Roy 2020).

The need for a method to rate the quality/relevance of questions is very important as the volume of questions on StackOverflow and such websites is large. If users become overwhelmed with a large volume of questions, the usage of services on those websites will be invalidated.

Those websites should eventually employ methods to (1) rate responses to questions and (2) rate the quality and relevance of original questions. Those ratings can be employed based on a mixture approach of (1) human-based annotation and (2) machine-learning based approaches. Our effort in this paper is related to the second approach. Specifically, we aim to apply neural network models to predict the quality of questions posted on StackOverflow website and evaluate their performance against traditional text mining classification models.

The remainder of the paper is structured as follows: section 2 offers an overview of existing literature pertaining to using deep learning with Question- Answering (QA) problems. Section 3, the methodology and dataset, explains the dataset, text preprocessing, and machine learning models developed. Section 4 shows the results of the machine learning models. The paper closes with section 5, which provides a summary of contributions and limitations.

## Literature review

According to the literature, several studies have addressed the issues on questions qualities on online question answering platforms.

Tóth et al., (2019) have presented an approach for classifying StackOverflow questions using the questions linguistic characteristics to automatically detect low quality questions posted by the users. Using natural language processing in combination with deep learning techniques, the authors trained a model using Nesterov Stochastic Gradient Descent method with momentum and evaluated the quality of classification using a test data set. The authors model achieved an accuracy of 74% in determining whether a question should be closed or not.

Menzies et al., (2018) evaluated alternative methods to deep learning on a dataset extracted from StackOverflow. They showed that simple alternative methods, based on tuning of clustering and individual classifiers can surpass the performance, efficiency, and speed of state-of-the-art deep learning models. According to the authors and in terms of performance and accuracy, those simple models perform as good as more complex deep learning models.

Hu et al., (2018) have proposed a "set of novel temporal features" as well as a deep learning framework for classifying and determining high quality questions posted in question answering Websites. A collaborative decision convolutional neural network to learn a set of features embedded in the collected data using different word embedding techniques was used. The authors also used non-linear semantic features to calculate a quality score for an answer. Finally, they conducted extensive experiments to show the advantages of the proposed framework. Experimentation and results showed that the proposed framework effectively predicts questions quality on question answering Websites.

Ruseti et al., (2018) assessed a number of recurrent neural networks architecture to predict question quality posted in question answering Websites. Data was collected and manually coded by raters based on corresponding depth and classified the data into four categories ranging from very shallow to very deep. The experiment evaluated different word embeddings. The experiments evaluated different recurrent neural networks architectures using "GRU, BiGRU and LSTM cell types of different sizes, and different word embeddings". Results showed that the best model achieved an accuracy of 81.22% which outperformed exiting results according to the literature.

Sen et al., (2020) have proposed an approach to address quality of question answering by utilizing a deep learning model using bidirectional transformers. The authors tested the applicability of the proposed approach using "bidirectional encoder representations from transformers trained on separate tasks". Results showed that increasing the pre-training of bidirectional encoder representations from transformers model as well as finetuning the question and answers can help improve the performance quality prediction and achieve a prediction accuracy higher than 80%.

Ahmadi, (2020) proposed a new method using deep neural networks for classifying project management related questions posted on question answering Websites. More than five thousand questions were collected and preprocessed then fed into a multi-input multi-head network. Data is received by the network separately and embedded internally, then extract the desired features, and complete the classification process. The proposed model was tested and compared with another four classifiers, namely, Random Forest, Naive Bayes, and SVM. Experimentation and results showed that the proposed classifier slightly outperformed the other four classifiers.

Zhang et al., (2022) proposed CCBERT, a deep learning based novel model to enhance the performance of the generation of StackOverflow question title by capturing the bi-modal semantic information from the entire question body and parsing the long-range dependencies to achieve better performance.

Al-Ramahi & Alsmadi, (2021) and (2020) also used deep learning models to predict the quality of Quora insincere questions. They utilized meta-text features in classical models and word-embedding in deep learning models to improve the performance of the developed prediction models.

Gupta & Kumar, (2021) proposed a method for generating an enhanced "class-specific word vector" that improves the unique properties of words in a class to address "light polysemy problem" related to qualification of question. The authors used different methodologies for questions classification (Word2Vec, Word2Vec+TF-IDF, and Word2Vec+Class Vector) in addition to the proposed approach and compared the performance of three classifiers, namely, CNN, Bi-LSTM and ABBC models. The tests were done using questions obtained from TREC, Kaggle and Yahoo. Results showed that the proposed approach outperformed the other three approaches used for classifying questions.

Roy, (2020) proposed a multi-layer convolutional neural network in order to reduce the number of insincere questions on Qoura Website. The authors have used two different word embedding techniques, namely, Skipgram and Continuous Bag of Word model, which both were trained using GloVe embedding vector. A machine learning convolutional neural network was used with the question text as an input, and the model predicted whether the question is considered insincere or not without using manual feature engineering. Results from the experiments showed that the proposed multilayer convolutional neural network achieved an F1-score of 98% for the best case.

Gaire et al., (2019) have tested machine learning models for classifying insincere question posted on Quora Website. The authors utilized an existing data set that consists of 1.6 million labeled and unlabeled examples. The set of questions were preprocessed by removing special characters, spaces, numbers, and cleaning misspelled words. Several word embedding techniques including GloVe and Paragram were used. Three supervised models using Multinomial Naïve Bayes, K-nearest, and Logistic Regression were trained and tested. Another neural network (RNN) was developed, trained, and tested. Results showed that the neural network outperformed the supervised models and achieved an F1 score of 69.13%.

According to the literature, quality of content on question answering website is still an important issue to address, since quality could be only supported using the text content of the posts (Tóth et al. 2019). Another issue with question answering websites is the lack of ground truth that can be used to judge and evaluate new question or post on these websites (Liu et al. 2017). A third issue with question answering website has to do with multiple questions, written differently, and having different syntax, but still provide the desired solution. Such similar question could end up with poor experience of users looking for an answer to the desired question (Dhakal et al. 2018; Kumar et al. 2019). Finally, question answering websites are overloaded with many questions and answers. For example, StackOverflow has 8,000 new posts on a daily basis (Tóth et al. 2019). Such large set of information could end up with the user not being able to filter the information or even choose the one that could help answering their problem (Kumar et al. 2019; Tóth et al. 2019)

Accordingly, in this paper we attempt to utilize a neural network model to predict the quality of questions posted on StackOverflow website and evaluate their performance against traditional text mining classification models.

## Methodology and dataset

### Dataset

In our experiments, we used the StackOverflow questions dataset (Annamoradnejad et al. 2022). This dataset contains 60,000 questions from 2016-2020 that are classified into three categories, high quality (HQ) questions, low quality questions that were edited (LQ Edit), and low-quality questions that were closed (LQ Close).

### Text preprocessing

As text preprocessing, we performed the following steps:
- Remove stop words.
- Convert the questions into a bag of words.
- We then used a binary weighting scheme so that 1 was assigned for word i in document j if the word i exists in the document j (i.e., $w_{i,j}=1$), $w_{i,j}=0$ otherwise.

### Machine learning models

To predict the quality of questions, we employed deep neural network models (DNN) against the baseline of several classification models such Naive Bayes (NB), Support Vector Machine (SVM), and Decision Tree (DT) classification models. For the network models, we created two sequential models, one with three dense layers (input, hidden and output layers) (see Table 1) and the other model with two dense layers (see Table 2).

| Layer (Type) | Output Shape | Param # |
| --- | --- | --- |
| dense_12 (Dense) | (None, 10) | 1,997,950 |
| dense_13 (Dense) | (None, 10) | 110 |
| dense_14 (Dense) | (None, 3) | 33 |
| Total params: 1,998,093, Trainable params: 1,998,093, non-trainable params: 0 | | |

Table 1. Neural network model: "sequential" with hidden layer

| Layer (Type) | Output Shape | Param # |
| --- | --- | --- |
| dense (Dense) | (None, 10) | 1,997,950 |
| dense_1 (Dense) | (None, 3) | 33 |
| Total params: 1,997,983, Trainable params: 1,997,983, non-trainable params: 0 | | |

Table 2: Neural Network Model: "sequential" without hidden layer

For model 1, let $L_i = \{L_1, L_2, L_3\}$. The hidden layer in the model transforms the input data into a different output dimension as shown in the following equation:

$$L_i = y_i = \varphi(w_i \otimes x_i + b) L_i = y_i = \varphi(w_i \otimes x_i + b)$$

Where at each forward pass at layer $L_i$, the weight or the kernel parameter is represented with $w_i$, the input vector is represented with $x_i$, and the bias is represented with $b$.

The projection function to transform input to output is represented with $\otimes$. Lastly, $\varphi$ represents the different activation functions. In our model, we used 'ReLU' as the activation function for the hidden layer, 'sparse_categorical_crossentropy' as a loss function, and 'Sigmoid' as the activation function for the output layer. To train the model, we used 80% of the dataset as training data and divided it into training and validation data. We used 30 epochs in the training process.

### Model evaluation

To evaluate the neural network and baseline models, we used 20% of the dataset as testing data. The performance of the developed DNN model was compared with the other baseline classification models using the accuracy measure. Accuracy is the percentage of correct classifications as shown in the following equation.

Accuracy = ((TP+TN))/((TP+FP+TN+FN))

Where TP is the true positive, TN is the true negative, FP is the false positive, FN is the false negative.

## Results

Table 3 shows the results of the baseline models. We tried different classical classifiers such as decision tree and support vector machine without parameter optimization. We also applied the logistic regression model using 10 folds cross validation along with hyper parameter optimization (parameter tuning using grid search).

| Model | F1 | Precision | Recall | Accuracy |
|---|---|---|---|---|
| DT | 0.73 | 0.73 | 0.73 | 0.73 |
| SVM | 0.77 | 0.77 | 0.77 | 0.77 |
| NB | 0.68 | 0.69 | 0.71 | 0.69 |
| LR | 0.74 | 0.75 | 0.74 | 0.74 |

Table 3: Baseline models

Table 4 shows the accuracy obtained when applying the neural network models (Keras sequential models with 30 epochs), model 1 with three layers and model 2 with two layers. The results show better performance of neural network models against baseline models with 80% accuracy.

| Model | Accuracy |
|---|---|
| Model 1: Three layers | 0.79 |
| Model 2: Two layers | 0.80 |

Table 4: Neural Network Results

Figure 1 and Figure 2 show the training curves of loss and accuracy on the train and validation datasets of neural networks models (model 1 and 2 respectively). The learning curves plots show changes in learning performance over time in terms of experience. The curves could be used to diagnose the overfitting of models. For example, Figure 1, learning curves of model 1 shows overfitting as the plot of validation loss decreases to a point and begins increasing again.

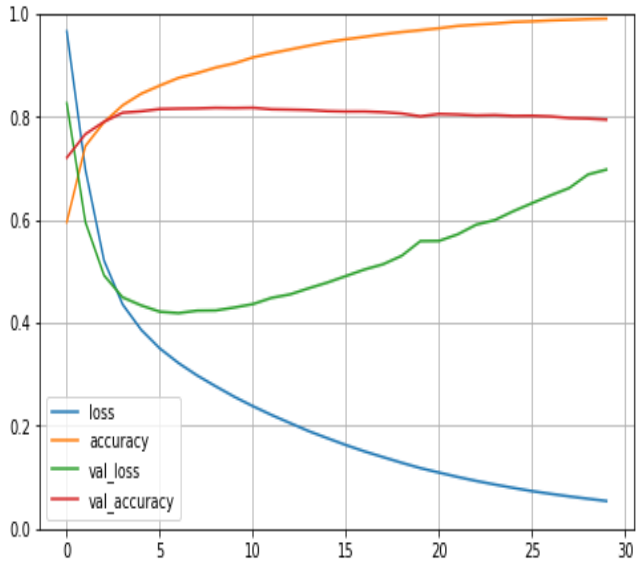

**Figure 1: The training and validation loss and accuracy curves of NN model 1**

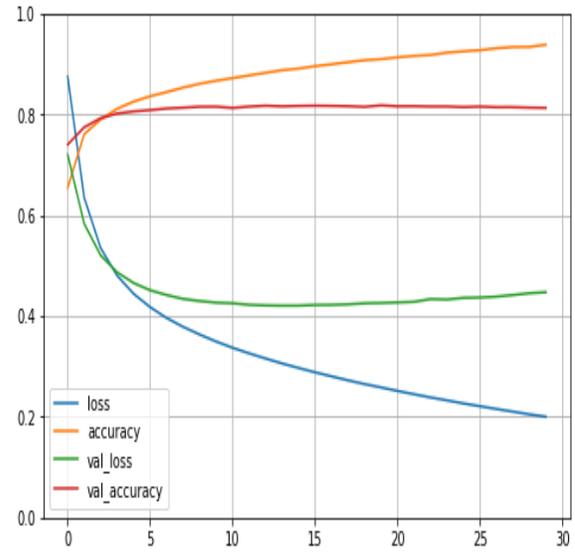

**Figure 2: The training and validation loss and accuracy curves of NN model 2**

## Discussion and conclusion

As volumes of information in social media websites is large, the ability of users to manually filter and evaluate quality in those websites is not realistic. As an alternative, many of those websites employ machine learning models to clean and filter irrelevant information created by users for different purposes. In this scope, we evaluated several machine learning models including neural networks to predict the quality of questions posted on StackOverflow website. Overall, results revealed better performance of the neural network models compared to traditional classification models. Therefore, demonstrate the efficacy of deep learning models in text classification over traditional text mining models. Further, the results show the performance of deep learning models can be affected by the deepness level of the network (i.e., the number of layers used). Hence, as future work, we plan to explore the optimization of neural network models to determine the optimal settings of the model (i.e., the number of hidden layers and neurons in each layer).

Overall, our study represents an advancement in the prediction of question quality on StackOverflow compared to existing research efforts. In contrast to the findings reported by Tóth et al. (2019), whose binary classifier achieved an accuracy of 74% in predicting question closure based on textual properties, our model shows a better performance. Specifically, our neural network model, Model 2, achieved an accuracy of 80% in discriminating between high-quality and low-quality questions. This demonstrates the effectiveness of the deep neural network model and dataset preprocessing techniques in capturing nuanced features indicative of question quality. Furthermore, our exploration of diverse machine learning models, including Support Vector Machine, Decision Tree, and Naive Bayes classifiers, further consolidates the robustness and generalizability of our predictive model. The utilization of our model and dataset could complement existing deep learning and natural language processing methods, contributing to the maintenance of high-quality questions at Q&A forums such as StackOverflow.

Our study is not without limitations, Figure 2 reveals a discrepancy in the accuracy between the training and validation datasets for Model 2, suggesting potential overfitting. This discrepancy raises concerns about the generalizability of our model to unseen data. As a result, to address this limitation, future research endeavors will involve the exploration of more advanced deep learning models incorporating regularization techniques. By incorporating regularization methods, we aim to mitigate overfitting and enhance the robustness of our predictive model.